\documentclass[letterpaper]{article} 
\usepackage[]{aaai23}  
\usepackage{times}  
\usepackage{helvet}  
\usepackage{courier}  
\usepackage[hyphens]{url}  
\usepackage{graphicx} 
\usepackage{natbib}  
\usepackage{caption} 
\usepackage[hidelinks]{hyperref}

\usepackage{todonotes}
\usepackage{amsmath,amsthm,amssymb}
\usepackage{thmtools,mathtools}

\usepackage{multirow}
\usepackage{booktabs}

\usepackage{algorithm}
\usepackage{algorithmicx}
\usepackage[noend]{algpseudocode}
\algrenewcommand\alglinenumber[1]{\scriptsize #1:}

\algnewcommand\algorithmicdatastructure[1]{\textbf{data structure}\ #1}
\algdef{SE}{DataStructure}{EndDataStructure}[1]{\algorithmicdatastructure{#1}}{\algorithmicend\ \algorithmicdatastructure}%

\makeatletter
\ifthenelse{\equal{\ALG@noend}{t}}%
  {\algtext*{EndDataStructure}}
  {}%
\makeatother
\MakeRobust{\Call}

\algnewcommand\True{\textbf{true}}
\algnewcommand\False{\textbf{false}}
\algnewcommand\Yield[1]{\textbf{yield}\ #1}
\newcommand{\algsetstretch}{\linespread{.9}\selectfont}

\makeatletter
\let\OldStatex\Statex
\renewcommand{\Statex}[1][3]{%
  \setlength\@tempdima{\algorithmicindent}%
  \OldStatex\hskip\dimexpr#1\@tempdima\relax}
\makeatother

\usepackage[noabbrev, capitalize]{cleveref}

\usepackage{tikz}
\usetikzlibrary{positioning, shapes, calc, chains, decorations.pathreplacing}

\usepackage{enumitem}
\setitemize{noitemsep,topsep=0pt,parsep=0pt,partopsep=0pt,leftmargin=0pt,itemindent=10pt}
\setenumerate{noitemsep,topsep=0pt,parsep=0pt,partopsep=0pt,leftmargin=0pt,itemindent=10pt}

\usepackage{xspace}
\usepackage{amsmath,amsthm,amssymb}

\newcommand{\tpset}{\ensuremath{\mathbb{T}}\xspace}

\newcommand{\deltastn}{\ensuremath{\operatorname{\delta-STN}}\xspace}
\newcommand{\deltaneighbors}{\ensuremath{\operatorname{\delta-Neighbors}}\xspace}
\newcommand{\clonestn}{\ensuremath{\operatorname{Clone-\deltastn}}\xspace}
\newcommand{\monostn}{\ensuremath{\operatorname{STN}}\xspace}
\newcommand{\smt}{\ensuremath{\operatorname{SMT}}\xspace}
\newcommand{\smtinc}{\ensuremath{\operatorname{SMT-Incremental}}\xspace}
\newcommand{\smtsua}{\ensuremath{\operatorname{SMT-SUA}}\xspace}
\newcommand{\linprog}{\ensuremath{\operatorname{LP}}\xspace}

\newcommand{\FCTP}{\ensuremath{\operatorname{FCTP}}\xspace}

\newcommand{\tuple}[1]{\langle #1 \rangle}

\newcommand{\var}[1]{\ensuremath{\text{\textit{#1}}\xspace}}

\newcommand{\optic}{\textsc{Optic}\xspace}
\newcommand{\tamer}{\textsc{Tamer}\xspace}

\newcommand{\painter}{\ensuremath{\textsc{Painter}}\xspace}
\newcommand{\Driverlog}{\ensuremath{\textsc{Driverlog}}\xspace}
\newcommand{\Satellite}{\ensuremath{\textsc{Satellite}}\xspace}
\newcommand{\TMS}{\ensuremath{\textsc{TMS}}\xspace}
\newcommand{\Floortile}{\ensuremath{\textsc{Floortile}}\xspace}
\newcommand{\MAP}{\ensuremath{\textsc{MapAnalyser}}\xspace}
\newcommand{\Matchcellar}{\ensuremath{\textsc{MatchCellar}}\xspace}
\newcommand{\majsp}{\ensuremath{\textsc{MAJSP}}\xspace}

\newtheorem{definition}{Definition}[section]

\title{An Efficient Incremental Simple Temporal Network Data Structure for Temporal Planning}
\author{Andrea Micheli}
\affiliations {
    Fondazione Bruno Kessler, Trento, Italy\\
    amicheli@fbk.eu
}

\nocopyright

\begin{document}

\maketitle

\begin{abstract}

One popular technique to solve temporal planning problems consists in decoupling the causal decisions, demanding them to heuristic search, from temporal decisions, demanding them to a simple temporal network (STN) solver. In this architecture, one needs to check the consistency of a series of STNs that are related one another, therefore having methods to incrementally re-use previous computations and that avoid expensive memory duplication is of paramount importance.

In this paper, we describe in detail how STNs are used in temporal planning, we identify a clear interface to support this use-case and we present an efficient data-structure implementing this interface that is both time- and memory-efficient. We show that our data structure, called \deltastn, is superior to other state-of-the-art approaches on temporal planning sequences of problems.

\end{abstract}

\section{Introduction}
\label{sec:intro}

Automated temporal planning concerns the synthesis of strategies to
reach a desired goal in a system subject to temporal constraints that
is specified by providing an initial condition together with
the possible actions that can drive it.

Over the years, several techniques to solve the temporal planning
problem have been presented; e.g. encoding into Satisfiability Modulo
Theory (SMT) \cite{shindavis}, plan-space planning \cite{europa} and
decision-epoch planning \cite{sapa}. One technique in particular
emerged to deal with temporally-expressive
\cite{cushing_temporally_expressive}, action-based temporal planning:
Forward-Chaining Temporal Planning (\FCTP) \cite{coles_stn}, embodied
in planners such as CRIKEY \cite{coles2008crikey,crikey-journal}, POPF
\cite{popf}, OPTIC \cite{optic}, VHPOP \cite{vhpop} and TAMER
\cite{tamer}. The basic idea behind \FCTP is to use forward heuristic
search to incrementally construct a causally-sound interleaving of
events and to check the temporal feasibility of any plan prefix using
scheduling techniques, where the timing of each event is encoded as a
symbolic variable.
%

When the planning problem at hand does not contain numeric variables
or continuous change (i.e. we are dealing with pure temporal
planning), the most efficient way to implement an \FCTP schema is to
use Simple Temporal Networks (STNs) \cite{stn} to represent and check
the consistency of the symbolic scheduling problems. STNs have been
widely studied and several algorithms to incrementally add and remove
constraints have been presented
\cite{cesta_oddi,ramaligam,gerevini,coles_stn}; moreover, STNs can be
easily encoded in Linear Programming (LP) or in SMT. However, all
these algorithms focus on how to efficiently solve, propagate and
maintain \emph{a single} STN; instead, in an \FCTP schema, we have one
STN associated to each search state, with minimal differences between
the STN of a state and the one of its predecessor in the search
(usually, each search step either introduces a new time-point in the
STN or adds precedence constraints).
%

In this paper, we present a novel data structure to efficiently
represent, maintain and solve STNs in an \FCTP schema. Our data
structure, called \deltastn, is persistent \cite{okasaki} and explicitly avoids the support for
retracting constraints (as this is never used in the \FCTP schema) and
efficiently saves memory, by maintaining references to STNs of
parent nodes in the search tree. Effectively, we identify the set of
operations needed for an STN to be employed in an \FCTP schema, we
show how to implement these in \deltastn and we show that \deltastn is
consistently superior in terms of both solving time and memory usage
to the other approaches and algorithms in the literature.

\section{STNs and their use in FCTP}

A Simple Temporal Network (STN) \cite{stn} is a data structure that
maintains constraints over a set of temporal variables called
time-points.
\begin{definition}
  An STN is a pair $\tuple{T, C}$ where $T$ is a set of time points
  and $C$ is a set of constraints of the form $x - y \le k$ with $x,y
  \in T$ and $k \in \mathbb{Q}$.
\end{definition}
An STN $\tuple{T, C}$ is said to be consistent if there exists an
assignment, called a consistent model, $\mu : T \rightarrow
\mathbb{Q}$ such that for each constraint $(x - y \le k) \in C$, the
inequality $\mu(x) - \mu(y) \le k$ is satisfied. Without loss of
generality, we assume that each $STN$ contains a reference zero time
point $z$ and a constraint $z - x \le 0$ for each time point $x$.

STNs are used in different areas of Artificial Intelligence to
maintain temporal knowledge and to efficiently check temporal
feasibility. In fact, STN consistency can be checked in polynomial
time by using a shortest path algorithm able to detect negative
cycles, such as the Bellman-Ford algorithm \cite{cormen}, over the
``Distance Graph'' (DG). The DG of a given STN is the graph having one
node for each time point, and an edge $\tuple{x, y}$ with weight $k$
for each constraint $y - x \le k$ in the STN. If the DG contains a
negative cycle, then the network is inconsistent; otherwise, for each
node $x$ we can set $\mu(x)$ to the distance from $x$ to $z$ to obtain
the earliest possible execution time for each time point (the
minimal-makespan consistent model) \cite{stn}.

We call ``Inverted Distance Graph'' (IDG) the graph where each edge
$\tuple{x, y}$ with weight $k$ of the DG is transformed into an edge
$\tuple{y, x}$ with weight $k$. It is easy to see that the IDG has a
negative cycle if and only if the DG has one (the weights do not
change and we can simply walk a negative cycle in the DG in reversed
order and vice-versa). Moreover, the shortest path distance from any
node $x$ to any node $y$ in the DG corresponds to the minimum distance
from node $y$ to node $x$ in the IDG. For convenience, in \deltastn
algorithms we will use the IDG and we will maintain and update the
distances from $z$ to each time point.

\begin{algorithm}[t]
\caption{\FCTP algorithm pseudocode}
\label{alg:fhsst}
\scriptsize
\algsetstretch
\begin{algorithmic}[1]
  \Procedure{FCTP}{\ }
    \State{$\var{init} \gets \tuple{\Call{InitialState}{\ }, \Call{MakeSTN}{\ }}$}\label{line:initstate}
    \State{$\var{Q} \gets \Call{HeuristicPriorityQueue}{}$; \ \ \Call{Push}{\var{Q}, \var{init}}}
    \While{\Call{NotEmpty}{Q}}
      \State{$\tuple{\var{c}, \var{cstn}} \gets \Call{Pop}{\var{Q}}$}
      \If{\Call{IsGoal}{\var{c}}} \label{line:goal}
         {\Return{\Call{GetPlan}{\var{c}, \Call{GetSTNModel}{\var{cstn}}}}}\label{line:mkplan}
      \Else
        \ForAll{$\var{child} \in \Call{CausalChildren}{\var{c}}$}
          \State{$\var{childstn} \gets \Call{CopySTN}{\var{cstn}}$} \label{line:stncopy}
          \ForAll{$\var{x} - \var{y} \le k \in \Call{TemporalConstraintsToAdd}{\ }$}
            \State{\Call{STNAdd}{\var{childstn}, \var{x}, \var{y}, k}} \label{line:stnadd}
          \EndFor
          \If{\Call{CheckSTN}{\var{childstn}}} \label{line:stncheck}
             {\Call{Push}{\var{Q}, $\tuple{\var{child}, \var{childstn}}$}}
          \EndIf
        \EndFor
      \EndIf
    \EndWhile
    \State{\Return{``No plan exists''}}
  \EndProcedure
\end{algorithmic}
\end{algorithm}

We now present the minimal interface that an STN must implement to
serve the needs of \FCTP. The general high-level schema of an \FCTP
planner is summarized in \cref{alg:fhsst}. Intuitively, classical
search states (i.e. total truth-value assignments) are augmented with
symbolic temporal information (in the form of a STN or an LP) encoding
the temporal constraints collected in the path so far. Starting from
an initial state where temporal constraints are empty
(\cref{line:initstate}), search states are expanded using a priority
queue (that uses some kind of heuristic ordering that is not relevant
for the purpose of this paper). Each time a state is expanded, either
it is recognized as a goal state (\cref{line:goal}) and a plan is
constructed from a consistent model of the STN (\cref{line:mkplan}) or
children are added to the queue. In the latter case, we construct all
the causally-sound children and for each of them we create an STN
inheriting all the constraints from the parent STN
(\cref{line:stncopy}). Moreover, we add the relevant temporal
constraints for the new state. Different algorithms add different
kinds of temporal constraints; for example, POPF \cite{popf}
constructs a partial order by analyzing supporters and deleters in the
prefix, while \tamer \cite{tamer} constructs a total
ordering. Finally, we check the resulting STN for consistency and we
only enqueue the children whose STN is consistent
(\cref{line:stncheck}).

This very high-level view of \FCTP highlights a simple, minimal
interface that is needed by an STN library to be used in this
context. Below, we describe each function of this interface
highlighting the underlying assumptions.

\begin{itemize}
\item \textbf{\textsc{MakeSTN}()} creates a new STN with no constraints.
\item \textbf{\textsc{CopySTN}($p$: STN)} creates a new STN, that
  inherits all the constraints of $p$.
\item \textbf{\textsc{STNAdd}($p$: STN, $x$: TimePoint, $y$:
  TimePoint, $b$: $\mathbb{Q}$)} adds a new constraint $x - y \le b$
  in the STN $p$. If needed, time points $x$ and $y$ are added in the
  set of time points. Multiple constraints can be added to an STN.
\item \textbf{\textsc{CheckSTN}($p$: STN)} returns true if the STN $p$ is
  consistent, false otherwise. If $p$ is consistent, the function
  internally stores a consistent model $\mu$.
\item \textbf{\textsc{GetSTNModel}($p$: STN, $x$:TimePoint)} Returns the
  value of the time-point $x$ in the consistent model $\mu$ stored by
  the \textsc{Check} function. This function, can only be called after
  a call to \textsc{Check} that returned true.
\end{itemize}

\section{\deltastn Algorithms}

We now describe our \deltastn data structure, implementing the defined
interface.
In the following, we indicate with $\tpset$ the set of all possible
time points. There are four key ideas that are at the base of the time
and memory efficiency of \deltastn. First, we leverage the intended use of the data structure in the \FCTP search schema by avoiding the duplicated representation of the constraints between a search node and its ancestors. In fact, we represent the set of temporal constraints in a \deltastn as an associative container (implemented with a sorted vector to save space and retain logarithmic access time), which maps a time point to a persistent list of pairs of a time point and a bound.
The second key idea stems from the observation that for the
incremental Bellman-Ford checking algorithm to work \cite{incbf}, we
only need to retrieve the constraints that are of the form $x - y \le
k$ given a specific time point $x$. We call such constraints
``neighbors'' of $x$. Our representation makes it easy to iterate through all the neighbors of $x$ as we can simply access the associative container in position $x$.
Third, we do not explicitly represent the reference time point
$z$: we implicitly impose the constraint $z - x \le 0$ for each time
point upon addition, setting the initial distance to $0$. Finally, we
copy the distances map in the \textsc{CopyTN} method, so that the
computation done in the parent STNs can be re-used in the children and
only additional constraints need to be incrementally propagated.

\begin{algorithm}[t]
\caption{\deltastn data structure}
\label{alg:datastructure}
\scriptsize
\algsetstretch
\begin{algorithmic}[1]
  \DataStructure{\deltaneighbors}
  \State{\var{dst} : \tpset}
  \State{\var{bound} : $\mathbb{Q}$}
  \State{\var{next} : pointer-to \deltaneighbors}
  \EndDataStructure

  \DataStructure{\deltastn}
  \State{\var{constraints} : map $\tpset \rightarrow$ pointer-to \deltaneighbors}
  \State{\var{distances} : map $\tpset \rightarrow \mathbb{Q}$}
  \State{\var{is\_sat} : $\mathbb{B}$}
  \EndDataStructure
\end{algorithmic}
\end{algorithm}

The data structure is reported in
\cref{alg:datastructure}. Essentially, the constraints of a \deltastn \var{stn} are represented as a pointer (\var{stn.constraints}) to a \deltaneighbors structure that is a functional linked list containing pairs of a time point and a bound (the empty list is indicated with $\bot$). This representation encodes a constraint $y - x \le k$ as an element $\tuple{y, k}$ in the list $\var{stn.constraints[x]}$. In this way, when an STN is copied to construct a descendant state in \FCTP, the constraints are not copied in memory, we simply copy the associative map retaining pointers to the \deltaneighbors list of the ancestor and new constraints are added as new heads of such list (see
\cref{line:shallowcopy} of \cref{alg:methods}).
Then, we have a map (distances)
from each time-point $x$ to the distance between $z$ and $x$ in the
IDG associated with the STN. The value of a time point $x$ in the
minimal-makespan STN model is just the $-1 \times $ distances[x] (see
\cref{line:invert} of \cref{alg:methods}). Finally, we incrementally
check and propagate each constraint upon addition, using the is\_sat
field to record if the network is consistent (is\_sat=true) or
not\footnote{By preventing constraint removal, once a \deltastn is
  declared not consistent (is\_sat=false), it is impossible to restore
  consistency.}. \deltastn stores the set of time points implicitly
in the set of keys of the distances map.

\begin{algorithm}[t]
\caption{\deltastn methods}
\label{alg:methods}
\scriptsize
\algsetstretch
\begin{algorithmic}[1]
  \Procedure{MakeSTN}{\ }
  \State{\textbf{return} \deltastn(\var{constraints}=$\emptyset$,
                                   \var{distances}=$\emptyset$,
                                   \var{is\_sat}=\True)}
  \EndProcedure

  \vskip 3pt
  \Procedure{CopySTN}{\var{p} : \deltastn}
  \State{\textbf{return} \deltastn(\var{constraints}=\Call{ShallowCopy}{\var{p.constraints}},} \label{line:shallowcopy}
    \Statex{\hskip27pt \var{distances}=\Call{Copy}{\var{p.distances}}, \var{is\_sat}=\var{p.is\_sat})}
  \EndProcedure

  \vskip 3pt
  \Procedure{STNAdd}{\var{self} : \deltastn, \var{x} : \tpset, \var{y} : \tpset, \var{b} : $\mathbb{Q}$}
    \If{\var{self.is\_sat}} \Comment{If self is not consistent, no need for more constraints}
      \ForAll{$\var{p} \in \{\var{x}, \var{y}\}$}
        \If{$\var{p} \not \in \var{self.distances}$} {$\var{self.distances[p]} \gets 0$} \EndIf
        \If{$\var{p} \not \in \var{self.constraints}$} {$\var{self.constraints[p]} \gets \bot$} \EndIf
      \EndFor
      \If{$\neg$ \Call{IsSubsumed}{\var{self}, \var{x}, \var{y}, \var{b}}} \label{line:subcheck}
          \State{\var{self.constraints}[\var{x}]$ \gets $\deltaneighbors(\var{dst}=\var{b}, \var{bound}=\var{b},} \label{line:consadd}
          \Statex{\hskip106pt \var{next}=\var{self.constraints}[\var{x}])}
          \State{$\var{self.is\_sat} \gets \Call{IncCheck}{\var{self}, \var{x}, \var{y}, \var{b}}$} \label{line:callinccheck}
      \EndIf
    \EndIf
  \EndProcedure

  \vskip 3pt
  \Procedure{CheckSTN}{\var{self}: \deltastn}
    \State{\Return{\var{self.is\_sat}}}
  \EndProcedure

  \vskip 3pt
  \Procedure{GetSTNModel}{\var{self}: \deltastn, \var{x} : \tpset}
    \State{\Return{$-1 \times \var{self.distances}[\var{x}]$}} \label{line:invert}
  \EndProcedure

  \vskip 3pt
  \Procedure{IsSubsumed}{\var{self}: \deltastn, \var{x} : \tpset, \var{y} : \tpset, \var{b} : $\mathbb{Q}$}
    \State{$\var{neighbors} \gets \var{self.constraints}[\var{x}]$}
    \While{$\var{neighbors} \not = \bot$}
      \If{$\var{neighbors.dst} = \var{y}$}
        \State{\Return{$(\var{neighbors.bound} \le \var{b})$}}\label{line:ordermatters} \Comment{Neighbors have stricter bounds first}
      \EndIf
      \State{$\var{neighbors} \gets \var{neighbors.next}$}
    \EndWhile
    \State{\Return{\False}}
  \EndProcedure

  \vskip 3pt
  \Procedure{IncCheck}{\var{self}: \deltastn, \var{x} : \tpset, \var{y} : \tpset, \var{b} : $\mathbb{Q}$}
    \If{$\var{self.distances}[\var{x}] + \var{b} < \var{self.distances}[\var{y}]$}
      \State{$\var{self.distances}[\var{y}] \gets \var{self.distances}[\var{x}] + \var{b}$}
    \Else
      {\ \Return{\True}}
    \EndIf
    \State{$\var{Q} \gets \Call{Queue}{\ }$; \ \ \Call{Push}{\var{Q}, \var{y}}}
    \While{\Call{NotEmpty}{\var{Q}}}
      \State{$\var{c} \gets \Call{Pop}{\var{Q}}$; \ \ $\var{n} \gets \var{self.constraints}[\var{c}]$}
      \While{$\var{n} \not = \bot$}
        \If{$\var{self.distances}[\var{c}] + \var{n.bound} < \var{self.distances}[\var{n.dst}]$}
          \If{$\var{n.dst} = \var{y} \wedge \var{n.bound} = \var{b}$}
            {\ \Return{\False}}
          \EndIf
          \State{$\var{self.distances[n.dst]} \gets \var{self.distances[c]} + \var{n.bound}$}
          \State{\Call{Push}{\var{Q}, \var{n.dst}}}
        \EndIf
        \State{$\var{n} \gets \var{n.next}$}
      \EndWhile
    \EndWhile
    \State{\Return{\True}}
  \EndProcedure
\end{algorithmic}
\end{algorithm}

The implementation of the interface defined in the previous section is
reported in \cref{alg:methods}. The \textsc{MakeSTN} method, simply
constructs an empty \deltastn with no ancestor (is\_sat is set to true
because the STN is vacuously consistent).  The \textsc{CopySTN}
method, also creates a new \deltastn, but this time the constraints of the parent \deltastn are inherited by shallow copying the \var{p.constraints} map (meaning that the map is copied, but the pointers to the \deltaneighbors structures in the values are not duplicated in memory). The distances and is\_sat values are simply copied from the parent. In this way, subsequent calls updating the consistency or the distances
will not affect the parent, because the \deltaneighbors structures are persistent, but parent constraints are still reachable and new constraints can be added on top of the parent constraints.  The core of the algorithm is
encapsulated in the \textsc{STNAdd} method.  We first extend the
distances and constraints maps, if needed, by setting the new time points to distance 0
(because an unconstrained time-point can always be set to 0 in a
consistent model) and the constraints to empty lists if not already present. At this point, we check (\cref{line:subcheck}) if
the constraint $x - y \le b$ is already subsumed by other constraints
in this network. This check is performed by the \textsc{IsSubsumed}
method that checks if this \deltastn already has a
constraint $x - y \le b'$ with $b' \le b$.
One drawback of a list representation instead of an associative container is that in order to recognize subsumption we require an iteration through the list that is in general $O(E)$ where $E$ is the number of constraints in the \deltastn. However, we limit this problem by maintaining a partial ordering property on the list of neighbors of each time point $x$: if there are two elements $\tuple{y, k}$ and $\tuple{y, k'}$ in $stn.constraints[x]$ with $k' < k$, then $\tuple{y, k'}$ appears before $\tuple{y, k}$ in the list. This property is maintained by the fact that we never add subsumed constraints (\cref{line:subcheck}) and is exploited by the \textsc{IsSubsumed} function to early terminate as soon as the target time point \var{y} is encountered in the list (\cref{line:ordermatters}).

If the constraint being added is found to be subsumed, then it means that it is
weaker than an already present constraint and can be
discarded. Instead, if the new constraint is not subsumed, we add it
to the set of constraints of this \deltastn (\cref{line:consadd}) and
we launch an incremental Bellman-Ford check from this constraint by
calling the \textsc{IncCheck} function (\cref{line:callinccheck}).
The \textsc{IncCheck} function is an incremental application of the
Bellman-Ford algorithm from \cite{incbf}; note that this
implementation exploits the previously computed distances and
guarantees to store the minimal valid consistent model in the
distances assignment. The retrieval of the neighbors of a
time point $x$ (\cref{line:neighbors}) is implemented by simply iterating over the elements of \var{self.constraints[x]}. Finally, we highlight that \deltastn can be destroyed and memory can be freed (for example when a state is closed by the \FCTP search), but we have to maintain the \deltaneighbors structures still in use by the descendants. For this, we simply use a standard reference-counting mechanism to maintain
consistency\footnote{Our implementation uses C++
  \texttt{std::shared\_ptr}, but any other system such as a garbage collector would work  as well.}.

\begin{figure*}
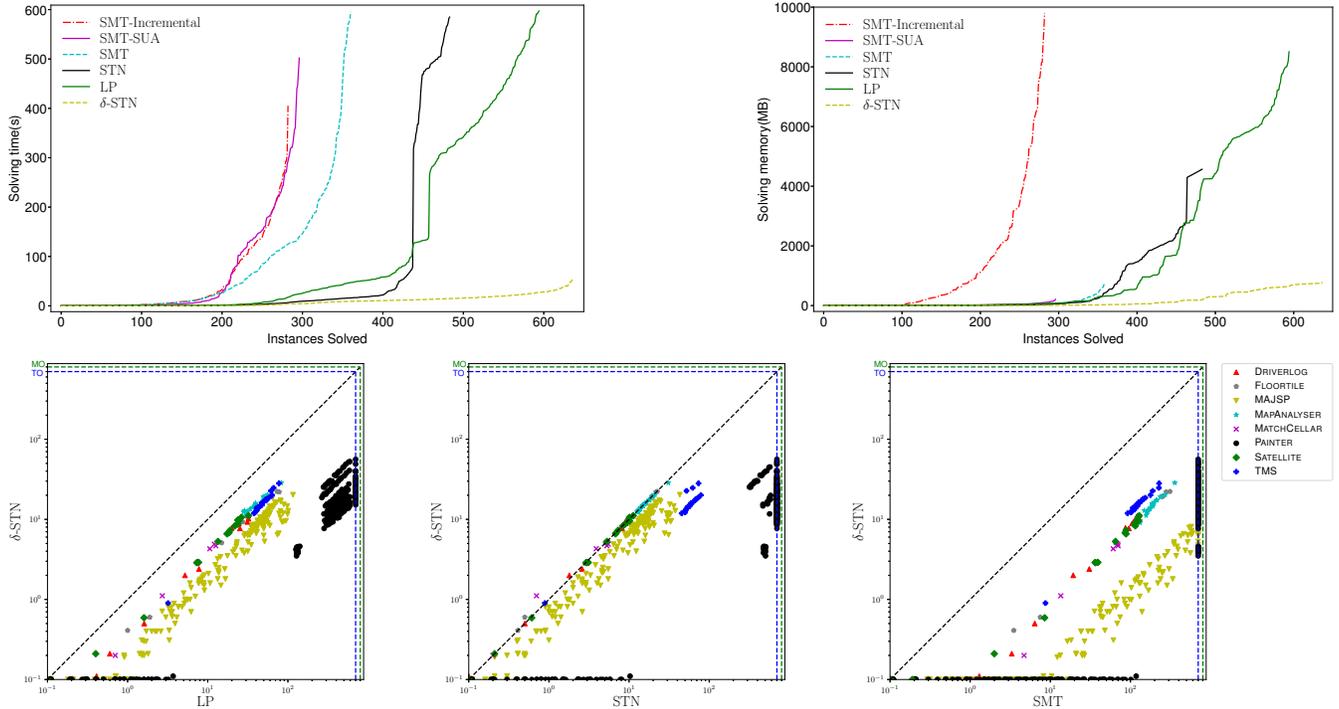

  \resizebox{.44\textwidth}{!}{\input{plots/cactus-time.pgf}}
  \hfill
  \resizebox{.44\textwidth}{!}{\input{plots/cactus-memory.pgf}}

  \resizebox{!}{.27\textwidth}{\input{plots/noleg-domain-scatter-lp-delta.pgf}}
  \hfill
  \resizebox{!}{.27\textwidth}{\input{plots/noleg-domain-scatter-mono-delta.pgf}}
  \hfill
  \resizebox{!}{.27\textwidth}{\input{plots/domain-scatter-smtmono-delta.pgf}}


  \caption{Experimental results. The top cactus plots show the overall
    time and memory comparison for all the approaches. The scatter
    plots below compare the time performance of the three best solvers
    with \deltastn.}
  \label{fig:results}
\end{figure*}

\section{Experimental Evaluation}
\label{sec:experiments}

In this section, we experimentally compare the merits of \deltastn
against several alternative techniques: linear programming using the
CLP \cite{clp} library used by \optic (indicated as \linprog), an
implementation of the standard Bellman-Ford algorithm with an optimized constraint representation using a Red-Black tree (indicated as \monostn), and three
implementations of the interface using Satisfiability Modulo Theory:
using no incrementality and re-constructing the problem each time
(\smt); cloning the problem (\smtinc) and using the ``solving under
assumptions'' interface (\smtsua).

We built 638 benchmark instances by instrumenting the \optic
and \tamer planners to output their STN/LP operations in textual
format. Each instance is a sequence of operations of the identified
interface extracted from a planning run. We let both the planners run
for 600 seconds on temporal domains from the IPC (\optic) and from the
domains used in \cite{tamer} (\tamer).

All the experiments have been conducted on a Xeon E5-2670 2.30GHz
machine with a runtime limit of 600 seconds and a memory limit of
10GB.
The source code implementation of \deltastn and of all the other
approaches is available for inspection in the additional material with
some example benchmarks and a Linux-64 binary to replicate the
results. The full set of benchmarks is available online
at \url{https://es-static.fbk.eu/people/amicheli/resources/deltastn}.

The results of the experimentation is reported in
\cref{fig:results}. The cactus plots in the top part of the figure
show the time and memory performance of the approaches in terms of the
number of solved instances. It is evident that \deltastn dominates
both the plots and it is the only implementation that is able to solve
all the 638 instances. The \linprog implementation is competitive but
falls quite far beyond \deltastn, solving 595 instances in total. The
\smt approaches are clearly not well-suited for this kind of operation
and are unable to cope with the high number of calls. In fact the \smt
interfaces are optimized for a ``linear'' use with the addition of
constraints, but poorly support branching a problem in different child
problems. This is also highlighted from the cactus plots in the bottom
part of the figure comparing the runtime of different approaches
against \deltastn. An interesting observation comes from the
comparison with \monostn: some instances are solved essentially in the same amount of time by duplicating the constraints in
each search state. This is
due to the fact that \optic does not internally use incrementality in
the way it interacts with CLP, hence some of the advantages of \deltastn
are not fully exploited.

\section{Conclusions}
\label{sec:conclusion}

We presented an optimized data structure to
solve Simple Temporal Networks in the \FCTP use-case. \deltastn
exploits the \FCTP characteristics to be more time- and
memory-efficient than other implementations.
In the future, we plan to study the use of \deltastn in other contexts
such as Partial Order Causal Planning \cite{traverso-book}.

\newpage

\section*{Acknowledgements}

I am very thankful to the anonymous reviewers of the ICAPS 2021 conference for the invaluable comments on a previous version of this manuscript.

Andrea Micheli has been partially supported by AIPlan4EU, a
project funded by EU Horizon 2020 research and innovation programme under GA n.\ 101016442.

\bibliography{refs.bib}

\begin{thebibliography}{20}
\providecommand{\natexlab}[1]{#1}

\bibitem[{Benton, Coles, and Coles(2012)}]{optic}
Benton, J.; Coles, A.~J.; and Coles, A. 2012.
\newblock Temporal Planning with Preferences and Time-Dependent Continuous
  Costs.
\newblock In \emph{{ICAPS} 2012}.

\bibitem[{Cesta and Oddi(1996)}]{cesta_oddi}
Cesta, A.; and Oddi, A. 1996.
\newblock Gaining Efficiency and Flexibility in the Simple Temporal Problem.
\newblock In \emph{Proceedings of the Third International Workshop on Temporal
  Representation and Reasoning, TIME-96, Key West, Florida, USA, May 19-20,
  1996}.

\bibitem[{Coles et~al.(2009{\natexlab{a}})Coles, Coles, Fox, and
  Long}]{coles_stn}
Coles, A.; Coles, A.; Fox, M.; and Long, D. 2009{\natexlab{a}}.
\newblock Incremental Constraint-Posting Algorithms in Interleaved Planning and
  Scheduling.
\newblock In \emph{Proceedings of the Workshop on Constraint Satisfaction
  Techniques for Planning and Scheduling (COPLAS '09), ICAPS 09}, 1 -- 8.
  Unknown Publisher.
\newblock COPLAS '09 ; Conference date: 01-01-2009.

\bibitem[{Coles et~al.(2009{\natexlab{b}})Coles, Fox, Halsey, Long, and
  Smith}]{crikey-journal}
Coles, A.; Fox, M.; Halsey, K.; Long, D.; and Smith, A. 2009{\natexlab{b}}.
\newblock Managing concurrency in temporal planning using planner-scheduler
  interaction.
\newblock \emph{Artif. Intell.}, 173(1): 1--44.

\bibitem[{Coles et~al.(2008)Coles, Fox, Long, and Smith}]{coles2008crikey}
Coles, A.; Fox, M.; Long, D.; and Smith, A. 2008.
\newblock Planning with Problems Requiring Temporal Coordination.
\newblock In \emph{AAAI}.

\bibitem[{Coles et~al.(2010)Coles, Coles, Fox, and Long}]{popf}
Coles, A.~J.; Coles, A.; Fox, M.; and Long, D. 2010.
\newblock Forward-Chaining Partial-Order Planning.
\newblock In \emph{{ICAPS} 2010}.

\bibitem[{Cormen et~al.(2001)Cormen, Stein, Rivest, and Leiserson}]{cormen}
Cormen, T.~H.; Stein, C.; Rivest, R.~L.; and Leiserson, C.~E. 2001.
\newblock \emph{Introduction to Algorithms}.
\newblock McGraw-Hill Higher Education, 2nd edition.
\newblock ISBN 0070131511.

\bibitem[{Cushing et~al.(2007)Cushing, Kambhampati, Mausam, and
  Weld}]{cushing_temporally_expressive}
Cushing, W.; Kambhampati, S.; Mausam; and Weld, D.~S. 2007.
\newblock When is Temporal Planning Really Temporal?
\newblock In \emph{{IJCAI} 2007}.

\bibitem[{Dechter, Meiri, and Pearl(1991)}]{stn}
Dechter, R.; Meiri, I.; and Pearl, J. 1991.
\newblock Temporal constraint networks.
\newblock \emph{Artificial intelligence}.

\bibitem[{Do and Kambhampati(2003)}]{sapa}
Do, M.~B.; and Kambhampati, S. 2003.
\newblock Sapa: {A} Multi-objective Metric Temporal Planner.
\newblock \emph{JAIR}.

\bibitem[{Frank and J{\'o}nsson(2003)}]{europa}
Frank, J.; and J{\'o}nsson, A. 2003.
\newblock Constraint-based attribute and interval planning.
\newblock \emph{Constraints}.

\bibitem[{Gerevini, Perini, and Ricci(1996)}]{gerevini}
Gerevini, A.; Perini, A.; and Ricci, F. 1996.
\newblock Incremental Algorithms for Managing Temporal Constraints.
\newblock In \emph{Eigth International Conference on Tools with Artificial
  Intelligence, {ICTAI} '96, Toulouse, France, November 16-19, 1996}, 360--365.
  {IEEE} Computer Society.

\bibitem[{Ghallab, Nau, and Traverso(2004)}]{traverso-book}
Ghallab, M.; Nau, D.~S.; and Traverso, P. 2004.
\newblock \emph{Automated planning - theory and practice}.
\newblock Elsevier.
\newblock ISBN 978-1-55860-856-6.

\bibitem[{johnjforrest et~al.(2019)johnjforrest, Vigerske, Ralphs, Hafer,
  jpfasano, Santos, Saltzman, h-i gassmann, Kristjansson, and King}]{clp}
johnjforrest; Vigerske, S.; Ralphs, T.; Hafer, L.; jpfasano; Santos, H.~G.;
  Saltzman, M.; h-i gassmann; Kristjansson, B.; and King, A. 2019.
\newblock coin-or/Clp: Version 1.17.3.

\bibitem[{Okasaki(1999)}]{okasaki}
Okasaki, C. 1999.
\newblock \emph{Purely functional data structures}.
\newblock Cambridge University Press.
\newblock ISBN 978-0-521-66350-2.

\bibitem[{Ramalingam and Reps(1996)}]{ramaligam}
Ramalingam, G.; and Reps, T.~W. 1996.
\newblock An Incremental Algorithm for a Generalization of the Shortest-Path
  Problem.
\newblock \emph{J. Algorithms}, 21(2): 267--305.

\bibitem[{Shin and Davis(2005)}]{shindavis}
Shin, J.-A.; and Davis, E. 2005.
\newblock Processes and continuous change in a SAT-based planner.
\newblock \emph{Artificial Intelligence}.

\bibitem[{Valentini, Micheli, and Cimatti(2020)}]{tamer}
Valentini, A.; Micheli, A.; and Cimatti, A. 2020.
\newblock Temporal Planning with Intermediate Conditions and Effects.
\newblock In \emph{AAAI 2020}.

\bibitem[{Wang et~al.(2005)Wang, Ivancic, Ganai, and Gupta}]{incbf}
Wang, C.; Ivancic, F.; Ganai, M.~K.; and Gupta, A. 2005.
\newblock Deciding Separation Logic Formulae by {SAT} and Incremental Negative
  Cycle Elimination.
\newblock In Sutcliffe, G.; and Voronkov, A., eds., \emph{Logic for
  Programming, Artificial Intelligence, and Reasoning, 12th International
  Conference, {LPAR} 2005, Montego Bay, Jamaica, December 2-6, 2005,
  Proceedings}, volume 3835 of \emph{Lecture Notes in Computer Science},
  322--336. Springer.

\bibitem[{Younes and Simmons(2003)}]{vhpop}
Younes, H.~L.; and Simmons, R.~G. 2003.
\newblock VHPOP: Versatile heuristic partial order planner.
\newblock \emph{Journal of Artificial Intelligence Research}.

\end{thebibliography}

\end{document}